\documentclass[letterpaper, 10 pt, conference]{ieeeconf}  

\IEEEoverridecommandlockouts                              

\usepackage{graphicx} 

\title{\LARGE \bf
Tactile Sensing Array for Multi-Phalanx Sensing in Humanoid Hands
}

\author{Neel Adwani, Muhaiminul Islam Akash, Rituja Bhattacharya, Cong Wang
\thanks{The authors are with New Jersey Institute of Technology (NJIT), Electrical and Computer Engineering, {\tt\small \{na657, ma2693, rb793, wangcong\}@njit.edu}}
}

\begin{document}

\maketitle
\thispagestyle{empty}
\pagestyle{empty}

\begin{abstract}



Humanoid hands require tactile feedback across the whole finger, not just the fingertip, to grasp and manipulate objects properly. 
Vision and proprioception alone cannot reliably provide this information, particularly when the hand's own fingers occlude the camera's view of the grasp.
We present a low-cost tactile array for a humanoid finger, made from Velostat and conductive tape. 
The fingertip carries seven contact points including a 2×3 matrix wrapped across its front, left, and right faces, and a separate contact point at the tip.
The proximal and middle phalanges each carry a single front-facing contact line. 
We measure the sensor's hysteresis and recovery time after release, through repeated loading and press-release tests.
We also test a compliant, 3D-printed contact structure with a gap and a bump, inspired by similar designs in prior work, and show it cuts recovery time by 74\% compared to a flush-contact baseline. 
We then show the sensor can produce distinct activation patterns for different contact geometries (flat, edge, corner) at the fingertip, and that it registers contact across all three phalanges during a grip. 
Finally, we discuss the limits of our fabrication changes and point to software-based compensation as a promising way to more directly fix the remaining hysteresis in the future.

\end{abstract}

\section{INTRODUCTION}
\label{sec introduction}


Domestic robots must handle objects with unknown weight, shape, friction, or deformability in unstructured, human-centric environments.
Doing this reliably takes more than open-loop grasping and manipulation using pre-planned trajectories \cite{kappassov2015tactile,dahiya2013tactile}.
It requires closed-loop control, which includes regulating grip force, localizing contact, detecting slip, and reacting to unexpected contact, all using tactile feedback.
In an unknown environment, vision alone is not enough because the hand's own fingers frequently block the camera's view of the grasp, and lighting conditions can add further uncertainty.
Proprioception is also insufficient, since joint angles alone reveal nothing about whether or how hard the hand is touching something.
The human hand compensates for these limitations through dense, distributed tactile feedback, which allows it to continuously adjust its grip during a task.
As humanoid hands are getting more sophisticated, equipping them with a comparable level of tactile sensing remains an open problem.


Tactile sensitivity, however, is not uniform across the hand.
The fingertip demands the richest and tri-directional sensing, while frontal sensing is adequate for proximal regions of the finger \cite{johansson1979tactile, almeida2025role}.
This has a direct design implication for fabrication: sensing every phalanx at the same high resolution wastes wiring and effort where it isn't needed.
Insufficient sensing of the fingertip, however, risks losing the contact information needed for precise manipulation.
Even at the fingertip itself, most existing tactile sensors provide only front-facing coverage and omit the lateral surfaces \cite{jia2026feel}, despite the fingertip routinely contacting objects along its sides during in-hand manipulation and power grasps. 
Section \ref{sec design} develops this distinction further and shows how it shaped our electrode placement.


Prior design in \cite{gomes2020geltip} achieves all-around coverage of the distal phalanx using a camera and optical housing, but this design is difficult to replicate at the scale of individual phalanges, making it impractical and expensive for multi-phalanx sensing.
Among available sensing materials, low-cost piezoresistive films such as Velostat are appealing.
They require no fluid injection or optical components, and are easy to wire and read out, making the tactile sensor easy to fabricate.
They are well-suited for covering the phalanges and the fingertip, giving us the flexibility to fabricate as many contact points as we need.
However, Velostat is known to exhibit significant hysteresis, drift, and slow post-contact relaxation, all of which introduce inconsistency in its readings.

In this work, we present a low-cost, Velostat matrix-electrode tactile array for the fingertip of a dexterous humanoid finger, with front-only sensing extended to the proximal and middle phalanges, motivated by the region-importance findings discussed above. 
This sensor can be developed by anyone with a hobby 3-D printer and a soldering station.
We characterize the sensor's hysteresis and post-release relaxation behavior through controlled loading and unloading, as well as repeated press-release testing. 
We further evaluate a compliant, 3D-printed gap-and-bump contact structure, inspired by contact-area-modulation designs in prior work, and report its effect on recovery time relative to a flush-contact baseline. 
Finally, we discuss the limitations of our fabrication-level interventions and identify software-based compensation, including simple exponential decay modeling of post-release relaxation, as a promising direction for more directly correcting residual hysteresis in future work.

\section{RELATED WORK}

Given the need for front-and-lateral, multi-phalanx sensing, realizing it in practice requires choosing a sensing technology that can be fabricated and deployed cheaply and repeatedly across many points on a single finger. High-fidelity tactile sensors based on vision, such as GelSight \cite{yuan2017gelsight} and DIGIT \cite{lambeta2020digit}, biomimetic sensors such as BioTac \cite{fishel2012biotac}, and magnetic skins such as uSkin \cite{tomo2016uskin} have a very high resolution, but typically trade off coverage against cost and integration complexity, and are often confined to the fingertip alone due to size and wiring constraints \cite{kappassov2015tactile}. More recently, conformable capacitive skins such as DexSkin \cite{wistreich2025dexskin} have achieved high-coverage tactile sensing across the finger surface and even the palm on platforms including the LEAP hand, but rely on more complex, application-specific fabrication than the simple resistive matrix approach explored here.

Tactile sensing in robotics is also achieved using Velostat, but it does not cover multiple phalanges with a curved surface and hysteresis behavior together. 3D-ViTac \cite{huang2024vitac} constructs a Velostat based tactile-sensor, covered with conductive thread matrix on fin-shaped gripper fingers, but does not characterize hysteresis or relaxation behavior. LeapTac \cite{jia2026feel} uses the same sensing principle to the front face of a fingertip of the LEAP Hand achieving a high resolution (7x6 grid), but reports only spatial resolution and minimum force threshold, without characterizing hysteresis or recovery time.

Separately, hysteresis in Velostat-based fingertip sensors has been addressed through hardware interventions. Prior work \cite{fang2022magnetic, bartunek2025layering} has reduced hysteresis-related error by pre-stressing Velostat using magnets to bias a single point and reduce non-linearity, and by stacking multiple Velostat layers, which can reduce measurement error by 27 - 60\%. However, this does not address recovery time, which we assume will increase because of stacking. A contact mechanism for Velostat-based tactile sensing \cite{zhang2024flexible} has achieved sub-100 ms response and recovery time using precision-molded soft electrodes and a gap/bump mechanism. None of this prior work evaluates low-cost, accessible fabrication interventions on a Velostat-based design involving multiple phalanges of a finger. We address this gap by testing a compliant, 3D printed spacer and bump contact structure on our own multi-contact point fingertip design, and report its effect on post-release recovery time.
We elaborate on these three gaps in the sensor design, testing and validation presented in the following sections.

\section{SENSOR DESIGN AND FABRICATION}
\label{sec design}

\subsection{Design Rationale}

\begin{figure}[t]
\centering
\includegraphics[width=0.75\columnwidth]{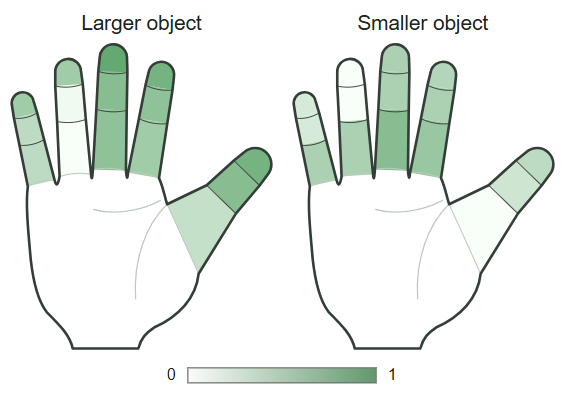}
\caption{Heatmap of important regions for grasping and manipulation, adapted from \cite{almeida2025role}}
\label{fig heatmap}
\end{figure}

Tactile sensing is not uniform across the hand.
The fingertip has the highest density of mechanoreceptors in the human hand, making it the primary region for tactile sensing \cite{johansson1979tactile}.
\cite{almeida2025role} confirms that once fingertip sensing is available, sensing on the middle and proximal phalanges becomes similarly important as they capture complementary information during grasping and manipulation.
They tested in-hand manipulation across different object shapes and sizes in simulation, and used the results to build a heatmap of which phalanges' sensing contributed the most to task performance, as shown in \ref{fig heatmap}.
Certain grasps such as large diameter and power sphere in the Cutkosky taxonomy \cite{cutkosky1989grasp} also require sensing on the middle and proximal phalanges.
Without this, the hand cannot distinguish a precision grasp from a prehensile one.

Grasping and manipulation contact in a fingertip occurs in palmar and lateral regions \cite{thomasson2022blind}.
Therefore, tactile sensing should extend to the front and sides of the finger, not just the fingertip pad, as most designs assume \cite{huang2024vitac, fishel2012biotac, jia2026feel}.
Beyond the front and sides, precision grasps tend to concentrate contact at the very tip of the finger, not across its broader surface. 
We account for this with a dedicated contact point at the distal tip, separate from the front and lateral sides, to directly capture contact at the point where fine manipulation and exploratory touch most often occur.

Beyond electrode placement, our contact mechanism is inspired by \cite{zhang2024flexible}, who report sub-100 ms response and recovery for a Velostat-based sensor using a gap-and-bump, contact-area-modulation structure. 
Their design relies on precision-molded soft electrodes and custom microfabrication, which is difficult to reproduce outside a specialized lab.
We realize the same underlying mechanism using accessible methods - i.e., standard 3D printing and off-the-shelf materials, so that the sensor can be fabricated without specialized equipment and therefore easily popularized.

\subsection{Electrode Layout and Signal Acquisition}

\begin{figure}[t]
\centering
\includegraphics[width=0.85\columnwidth]{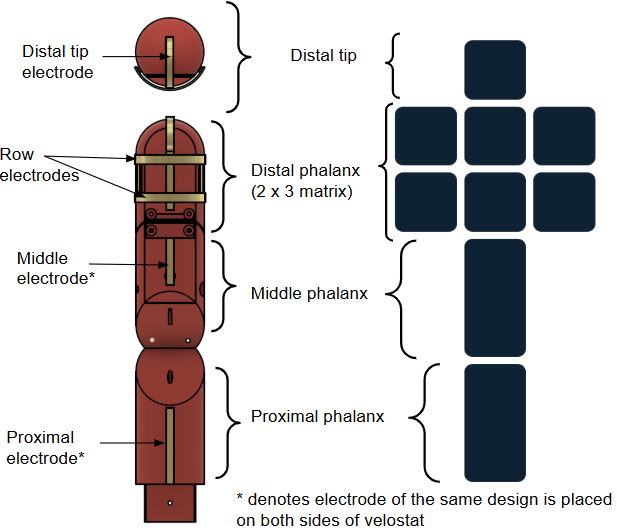}
\caption{Electrode Layout}
\label{fig electrode layout}
\end{figure}

The distal phalanx carries seven contact points in total. 
Six of them form a 2×3 row-column matrix wrapped around the phalanx's front, left, and right faces.
One column of contact points per face, sharing two row electrodes across six contact points from five electrical wires. 
A seventh contact point, wired independently, sits at the very tip of the distal phalanx, where the nail lies. 
The proximal and middle phalanges each carry a single front-facing contact line, with one electrode on either side of the piezoresistive layer. 
Fig. \ref{fig electrode layout} shows the full electrode layout, alongside the corresponding sensing-matrix visualization from our visualization GUI, in which each physical contact point maps to a colored block whose intensity reflects the sensed pressure at that location. 
This visualization is used in Section~\ref{sec testing results} to report results.

Velostat is used for the piezoresistive layer, with conductive tape as electrodes. 
The 2×3 matrix is scanned column by column for each row at 20 Hz. 
One row is driven HIGH at a time while the others float, and each column is read through a voltage-divider circuit with a pull-down resistor, sized empirically from Velostat's measured resistance range. 
The tip contact point and the proximal and middle contact lines use the same voltage divider principle, but without multiplexing, which is consistent with their role \cite{almeida2025role}.





\subsection{Fabrication Process}



The distal phalanx's sensing stack includes the following components:
\begin{enumerate}
    \item A backing plate,
    \item Row electrodes bonded to the backing plate,
    \item A Velostat layer,
    \item A spacer with seven holes for creating a deliberate air gap at each contact point,
    \item Column electrodes bonded to the spacer,
    \item A bump plate with a raised bump above each hole to concentrate contact force.
\end{enumerate}

\begin{figure}[t]
\centering
\includegraphics[width=0.85\columnwidth]{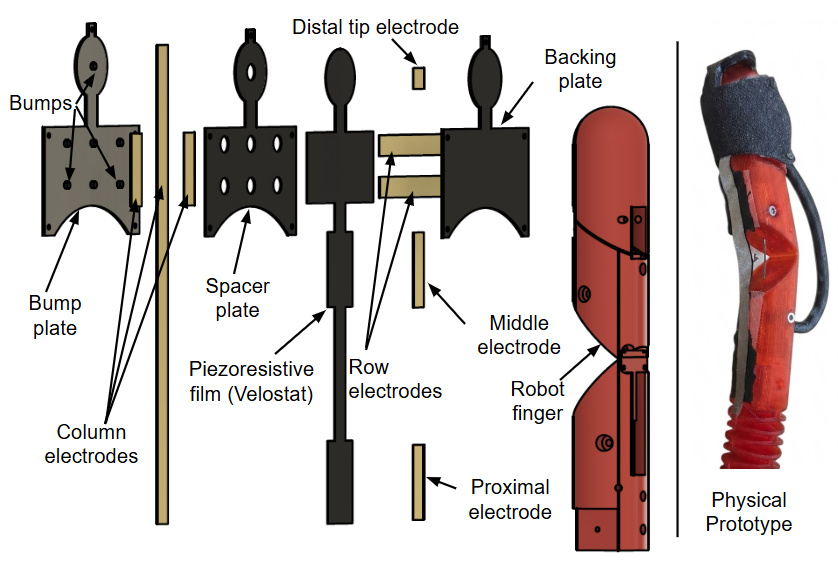}
\caption{Exploded view and prototype of the tactile sensor}
\label{fig sensor prototype}
\end{figure}

The backing, spacer, and bump plates were modeled in Fusion 360 and 3D-printed in TPU. 
Assembly is as follows: 
\begin{enumerate}
\item Two rows of conductive tape were placed equidistant on the TPU backing plate, along with the distal tip's electrode.
\item Velostat is double-taped over the non-electrode areas.
\item The spacer is placed on top of the Velostat.
\item Three columns of conductive tape were placed on the spacer, along with the tip's electrode.
\item The bump plate was added as the outermost layer, and all the layers were bolted together.
\end{enumerate}
The construction of the sensor for proximal and middle phalanges is relatively straightforward due to their simpler structure.
The parts and sensor stack are shown in Fig.~\ref{fig sensor prototype}.
We tested several spacer and bump widths and thicknesses, and selected the dimensions that gave the best recovery time, as detailed in Section~\ref{sec testing results}.

\section{TESTING AND RESULTS}
\label{sec testing results}

\subsection{Hysteresis Characterization}

\begin{figure}[t]
\centering
\includegraphics[width=0.85\columnwidth]{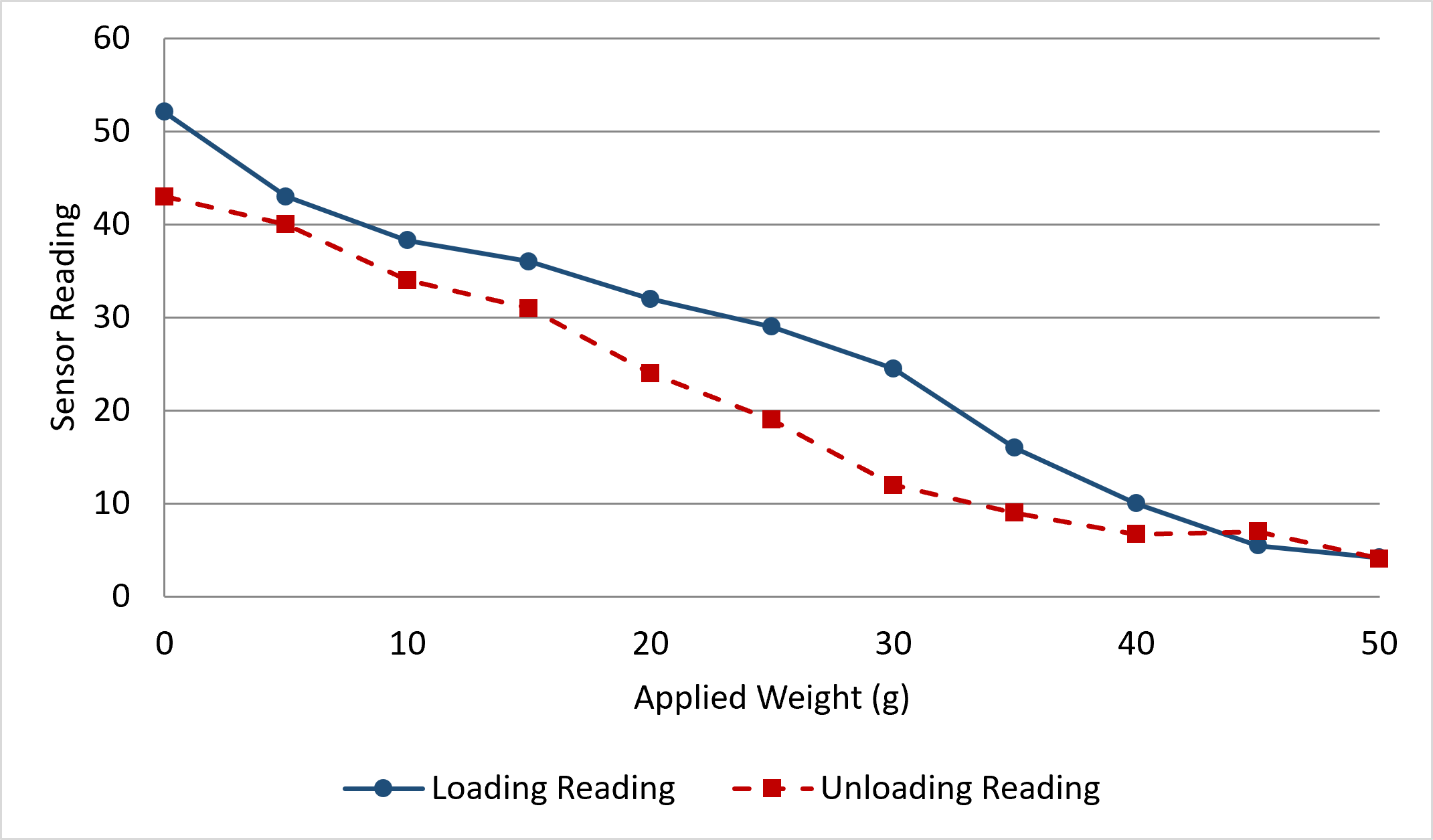}
\caption{Hysteresis curve of the proposed sensor}
\label{fig hysteresis}
\end{figure}

To characterize hysteresis, a single representative contact point was loaded in increments of 5 grams on a scale, allowing simultaneous readout of applied force and raw sensor value. 
Force was increased gradually from 0 g to 50 g, then decreased back to 0 g, with the raw ADC reading recorded at multiple points along both the loading and unloading paths. 
Fig.~\ref{fig hysteresis} shows the resulting loading and unloading curves. 
The two curves do not coincide, confirming the presence of hysteresis: at a given applied force, the reading differs depending on whether the sensor is being loaded or unloaded, consistent with the visco-elastic behavior reported for Velostat in prior work~\cite{zhang2024flexible, bartunek2025layering}.

\subsection{Post-Release Recovery Time}

To evaluate the effect of our gap-and-bump contact structure on recovery time, we compared a flush-contact baseline design against the compliant TPU gap-and-bump design described in Section \ref{sec design}-C. 
For each configuration, a known weight was applied to a single contact point, held briefly, and released, and the time for the raw signal to return to within 20\% of its pre-contact baseline was recorded over 5 repeated trials. 
The gap-and-bump design reduced mean recovery time from 4.6 ± 1.4 s to 1.2 ± 0.7 s, based on 5 trials, which is a 74\% reduction. 
The optimal spacer thickness was 0.4 mm, and the optimal bump height was 0.6 mm.
We note that our spacer and bump layers were both fabricated from TPU, rather than a rigid spacer paired with a compliant bump as in \cite{zhang2024flexible}.

\subsection{Contact Discrimination}

\begin{figure}[t]
\centering
\includegraphics[width=0.85\columnwidth]{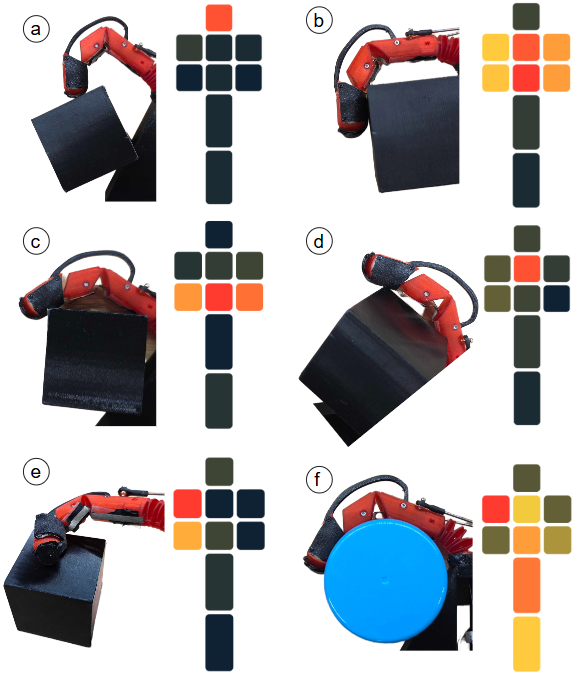}
\caption{Visualization of sensing matrices}
\label{fig contact discrimination}
\end{figure}

We evaluated the sensor stack on the distal phalanx's 2×3 wrapped matrix and tip contact point, together with the single proximal and middle phalanx contact lines in two scenarios. 
First, the assembled fingertip was pressed against a cube in five different ways: tip touching the surface, fingerpad touching the flat surface, edge, a corner, and laterally touching the flat surface. 
Fig.~\ref{fig contact discrimination} (a - e) shows the resulting activation across the seven distal-phalanx contact points for each orientation.
Contact was concentrated at the tip point alone for tip contact (a), broadly across the front of the matrix for frontal contact (b), along a line of points for edge contact (c), at a single point for corner contact (d), and across the right for lateral flat contact (e).
Second, the finger was used to grip a cylinder such that the distal, middle, and proximal phalanges made contact simultaneously, exercising the full sensing array. 
Fig. \ref{fig contact discrimination} (f) shows the resulting activation across all three phalanges during this grip, confirming that the middle and proximal contact lines register contact alongside the distal matrix as intended by the design rationale in Section \ref{sec design}.

\section{CONCLUSIONS AND FUTURE WORK}



We presented a low-cost, multi-phalanx Velostat tactile sensor for a humanoid finger. The design is based on evidence that different parts of the finger need different levels of sensing.
Richer, tri-directional sensing is required at the fingertip, and simpler, front-facing sensing at the proximal and middle phalanges. 
We measured the sensor's hysteresis and recovery time after release, and showed that a compliant, 3D-printed contact structure with a gap and a bump reduces recovery time compared to a flush-contact baseline. 
We also showed that the sensor can produce different patterns based on different geometries at the fingertip and registers contact across all three phalanges during a grip, matching what our design was meant to do.

Future work includes extending this sensor design across the whole hand, not just a single finger, so the palm and all fingers can share tactile feedback during a grasp. 
This would let us test ``blind'' grasping and manipulation, which will involve tasks done using tactile feedback alone, without vision to see how well the sensor supports real manipulation.
Additionally, we will be testing on a wider range of object shapes and sizes, beyond the cube and cylinder used in this work, to see how well the contact-discrimination results generalize.
We will be integrating the kinematic state of the finger (joint angle, joint velocity, and how long a contact has lasted) into the hysteresis correction, rather than relying on the tactile signal alone. 

\nocite{yuan2017gelsight}   
\bibliographystyle{IEEEtran}
\bibliography{refs}

\end{document}